\documentclass[10pt,twocolumn,letterpaper]{article}

\usepackage{cvpr}              

\usepackage[T1]{fontenc}
\usepackage{titling}
\usepackage{graphicx}
\usepackage{amsmath}
\usepackage{amssymb}
\usepackage{booktabs}
\usepackage{float}

\usepackage[pagebackref,breaklinks,colorlinks]{hyperref}

\usepackage[capitalize]{cleveref}
\crefname{section}{Sec.}{Secs.}
\Crefname{section}{Section}{Sections}
\Crefname{table}{Table}{Tables}
\crefname{table}{Tab.}{Tabs.}

\begin{document}
\setlength{\droptitle}{-5em}
\title{Investigating Fouling Efficiency in Football \\ Using Expected Booking (xB) Model}

\author{
Adnan Azmat \\
School of Computer Science and Engineering \\
Nanyang Technological University, Singapore\\
adnan002@e.ntu.edu.sg
\and
Su Su Yi \\
School of Computer Science and Engineering \\
Nanyang Technological University, Singapore\\
susu001@e.ntu.edu.sg
}
\predate{} \postdate{} \date{}
\maketitle

\begin{abstract}
   \noindent This paper introduces the Expected Booking (xB) model, a novel metric designed to estimate the likelihood of a foul resulting in a yellow card in football. Through three iterative experiments, employing ensemble methods, the model demonstrates improved performance with additional features and an expanded dataset. Analysis of FIFA World Cup 2022 data validates the model's efficacy in providing insights into team and player fouling tactics, aligning with actual defensive performance. The xB model addresses a gap in fouling efficiency examination, emphasizing defensive strategies which often overlooked. Further enhancements are suggested through the incorporation of comprehensive data and spatial features.
\end{abstract}

\section{INTRODUCTION}
\label{sec:intro}

\noindent In the realm of football, every decision made on the field can significantly influence the game's outcome. Among these decisions, fouls and their subsequent consequences are often overlooked, despite their pivotal role in the game. Current methodologies primarily focus on offensive actions leading to goals, leaving the subtler yet equally crucial defensive aspect under-emphasized. Analyzing defensive play poses inherent challenges, and football discussions often prioritize offensive actions, thereby overshadowing the importance of defensive contributions.

\subsection{Problem Statement}

\noindent In this paper, we introduce a new metric, the \textbf{Expected Booking (xB)} model. This model is designed to estimate the likelihood of a player receiving a yellow card if/for a foul committed at a specific moment during a match. Our focus is primarily on yellow cards due to their relative ambiguity and higher frequency of issuance compared to red cards. Red cards, which are issued less frequently, have clear rules for issuance. On average, yellow cards are issued 18 times more frequently than red cards \cite{poli2020global}.

\noindent \\ Our attention is more specifically directed towards non-dangerous fouls, excluding those resulting from bad behavior. We believe that non-dangerous fouls, excluding those resulting from bad behavior, truly represent the tactical aspect of fouls and do not involve impulsive decisions by players that easily lead to bookings.

\noindent \\ We anticipate that the cumulative Expected Booking probability can be used to rate the fouling ability of teams and players over individual matches or seasons, similar to how expected goals are used to rate scoring abilities. This approach offers a comprehensive view of the game, highlighting both offensive and defensive contributions. It also provides an opportunity to study the differences in refereeing decisions across various leagues and combinations.

\subsection{Motivation}

\noindent Most efforts in football analysis have been predominantly focused on aspects such as attacking, scoring, and passing. While statistics for offensive plays like goals scored, expected goals, assists, expected assists, dribbles, key passes, pass accuracy, and chances created are readily available and comprehensive, the metrics for defensive actions often lack depth and detail. 

\noindent \\ Defensive actions are typically represented by basic statistics like fouls, bookings (yellow cards, red cards), tackles, interceptions, and clearances. However, these metrics do not fully capture the complexity and strategic importance of defensive plays in a football match. This is the motivation behind the Expected Booking (xB) model.

\noindent \\ Inspired by the Expected Goals model, which provides a measure of a striker's performance relative to the opportunities they had, we aim to develop a similar model for defensive actions. To the best of our knowledge, there has been no prior research or study conducted on or related to the Expected Booking model. We believe that our research has the potential to offer valuable insights and contributions to the subject matter.

\subsection{Related Work}

\noindent \textbf{Expected Goals (xG) Model}. Rathke, A. 2017 examined goal scoring in European football leagues and factors which are associated with predicting Expected Goals (xG) \cite{rathke2017examination}. He analysed the factors, distance of the shot taken from goal and the angle of the shot in relation to the goal and used Expected Goals Method (xG) to calculate the chances a team has to score and concede goals \cite{rathke2017examination}. xG models use statistical and machine learning techniques to predict the probability of a shot resulting in a goal. These models often consider various features such as shot location, angle, and player positions. 

\noindent \\ \textbf{VAEP Framework}. T Decroos et al., 2019 proposed VAEP (Valuing Actions by Estimating Probabilities) framework for valuing actions performed by soccer players \cite{decroos2019actions}. The VAEP framework provides a simple approach to valuing actions that is independent of the representation used to describe the actions \cite{decroos2019actions}.  It considers all types of actions (e.g., passes, crosses, dribbles, take-ons, and shots) and accounts for the circumstances under which each of these actions happened as well as their possible longer-term effects \cite{decroos2019actions}. 

\noindent \\ \textbf{Expected Threat (xT) Model} . K Singh introduced xT (Expected Threat) model for team behaviour in possession to gain a deeper understanding of buildup play \cite{singh2021introducing}. K Singh proposed that each action should be assigned a score in isolation, disregarding what happened before and after it in the possession \cite{singh2021introducing}. One simplified way of viewing buildup play is as follows: when a team has possession in a certain position, they can either shoot (and score with some probability), or move the ball to a different location via a pass or a dribble \cite{singh2021introducing}. This continues until the team either loses possession, or scores a goal \cite{singh2021introducing}. xT model assigns a score to each player action (pass or dribble) based on how much it contributed to the buildup play \cite{singh2021introducing}. The point of xT is to come up with a metric that can quantify threat at any location on the pitch \cite{singh2021introducing}.


\section{DATA and FRAMEWORKS}

\noindent The data used in our research is primarily sourced from StatsBomb \cite{statsbomb}, which provides event data and 360 data. Football event data is a collection of detailed information about every event that occur during a football match. The data includes information about the location of the ball, the players’ movements, and the actions they take, such as passes, shots, and tackles. The data also includes information about the outcome of each event, such as whether a pass was successful or not.

\subsection{StatsBomb Event Data}

\noindent StatsBomb event data is renowned for its detail and accuracy. It records over 3,400 events per match, providing a comprehensive view of each game. This data includes unique features that allow for deeper analysis.

\subsection{StatsBomb 360 Data}

\noindent StatsBomb 360 is an extension to the football event data. It adds player locations to the event data, providing a more contextual view of the game. This data is collected using a hybrid of computer vision and manual oversight, ensuring its accuracy and reliability. However, it is important to note that the availability of open-source 360 data is significantly less compared to event data.

\subsection{Socceraction}

\noindent In addition to StatsBomb data, we also use Socceraction \cite{socceraction}, a Python package for objectively quantifying the impact of individual actions performed by soccer players using event stream data. The general idea is to assign a value to each on-the-ball action based on the action's impact on the game outcome, while accounting for the context in which the action happened. This allows us to measure the effectiveness of each player's actions in a game.

\section{FEATURES and LABELS}

\noindent We filter the event data specifically for fouling events. After the data pre-processing and preparation steps, each instance in our training data represents a non-dangerous foul event, which is not a result of bad behavior. Non-dangerous fouls are identified by excluding fouls due to handball, dangerous play, foul out, dive, 6 seconds, and backpass pick.

\subsection{Features}

\noindent In developing the Expected Booking (xB) model, we leveraged our football domain knowledge and employed statistical methods to identify key features for the model. For the purpose of this analysis, the team of the fouled player is referred to as the team in possession, and the fouling player’s team is referred to as the defending team.

\subsubsection{Minutes}
\noindent This feature represents the exact minute during the match when the foul was committed. It provides context about the stage of the game, which can influence the likelihood of a booking.

\subsubsection{Distance to Goal}
\noindent This feature measures the distance from the location of the foul event to the goal of the team in possession. This feature is represented by the black dotted line in figure 2.

\subsubsection{Angle to Goal}
\noindent This feature calculates the angle from the location of the foul event to the midpoint of the defending team’s goal. This feature is represented by the pink arc in figure 2.

\subsubsection{Foul Count Player}
\noindent This feature counts the number of fouls committed by the fouling player in the current match up to the time of the current foul. It serves as an indicator of the player's overall aggression or recklessness within the specific game.

\subsubsection{Foul Count Team}
\noindent This feature counts the number of fouls committed by the defending team in the current match up to the time of the current foul. It provides insight into the team's overall defensive strategy and discipline within the specific game.

\subsubsection{Goal Difference}
\noindent This feature calculates the goal difference between the team in possession and the defending team at the time of the foul. It can provide context about the competitive dynamics of the match.

\subsubsection{VAEP Offensive}
\noindent VAEP (Valuing Actions by Estimating
Probabilities) is a framework for valuing actions performed by players \cite{decroos2019actions}. VAEP value of an action is the sum of action’s offensive value and defensive value \cite{decroos2019actions}. It is given by the equation: \(V(a_i, x) = \Delta P_{\text{scores}}(a_i, x) + (-\Delta P_{\text{concedes}}(a_i, x))\).

\begin{figure}[H]
    \centering
    \includegraphics[width=1\linewidth, height=0.9\linewidth]{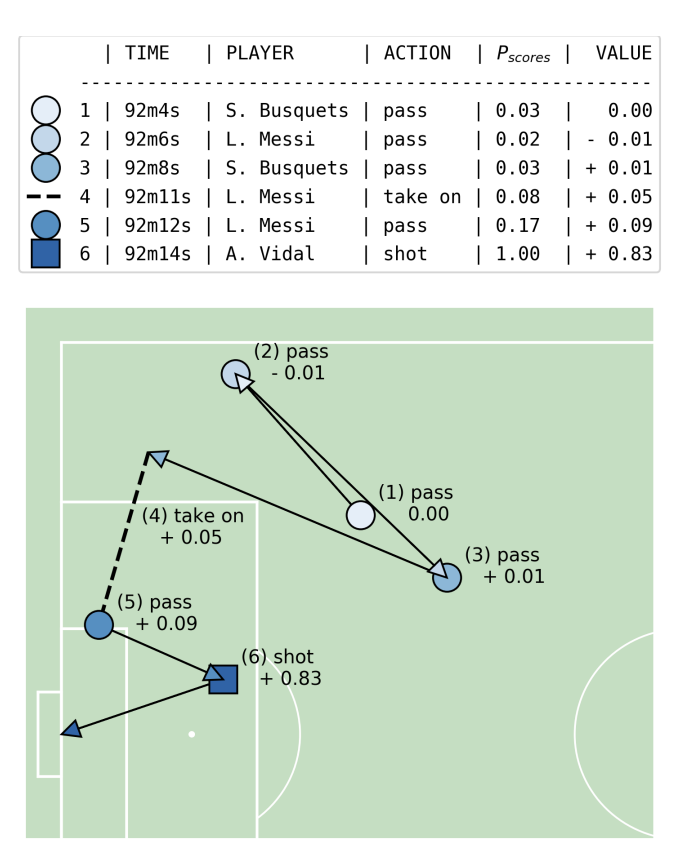}
    \caption{VAEP values for each action in a sequence of plays}
    \label{fig:enter-label}
\end{figure}

\noindent \\ VAEP offensive is denoted by $P_{scores}$. This feature captures the offensive threat posed by the team in possession \cite{decroos2019actions}. The rationale is that the defending team is more likely to receive a card if the team in possession poses a significant offensive threat \cite{decroos2019actions}.

\subsubsection{Features from 360 data}
\noindent We utilize the comprehensive StatsBomb 360 data to provide additional context to the game. Although the 360 data does not include the coordinates of all players on the field, it does provide a ‘freeze frames’ column which contains the coordinates of all players visible in the camera frame at the time of the foul.

\noindent \\ \textit{A. Attackers Count}
\noindent \\ \\This feature counts the number of players from the team in possession who are located ahead of the foul location. It provides insight into the attacking formation and strategy of the team in possession. This feature is represented by the red numberings: 1 and 2 in figure 2.

\noindent \\ \textit{B. Defenders Count} 
\noindent \\ \\ This feature counts the number of players from the defending team who are positioned between the foul location and their own goal. It provides insight into the defensive formation and strategy of the defending team. This feature is represented by the blue numberings: 1, 2 and 3 in figure 2.

\begin{figure}[H]
    \centering
    \includegraphics[width=1\linewidth]{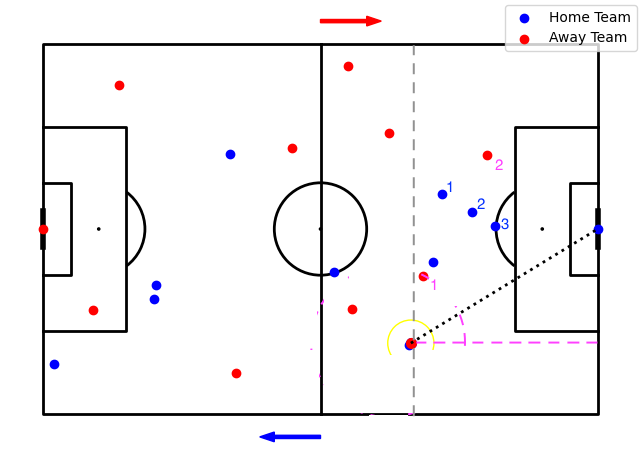}
    \caption{Dummy instance to show feature values}
    \label{fig:enter-label}
\end{figure}

\subsection{Classification Labels}

\noindent Our classification task involves binary outcomes, transforming it into a binary classification problem. Each feature sample is associated with one of two possibilities: either receiving a yellow card or not.

\noindent \\ However, it's worth noting that even though this is a binary classification task, we can additionally estimate probability of a booking (yellow card) for any given data instance. The cumulative probability of individual foul events offers insights into player behavior, team dynamics, and competition trends, enhancing our understanding of booking incidents.

\section{EXPERIMENT}
\noindent We conducted three iterations of experiments utilizing StatsBomb event and 360 data from various competitions.

\subsection{Naive Exploration: Preliminary Findings}
\noindent We leverage the limited 360 dataset comprising 957 foul events from 4 competitions and 210 matches, encompassing six features: minutes, score difference, distance to goal, angle to goal, foul count player and foul count team. The dataset is then partitioned into training and testing sets, following an 80-20 split. We proceed to train the model using two traditional classifiers, namely Decision Tree and Logistic Regression, alongside two ensemble classifiers—Gradient Boosting and XGBoost, ensuring a comprehensive exploration of diverse modeling techniques.

\begin{table}[!h]
\small 
\label{tab:my_table}
\begin{tabular}{|l|c|c|c|c|} \hline 
                                                                                     & \textbf{\begin{tabular}[c]{@{}c@{}}Decision\\ Tree\end{tabular}} & \textbf{\begin{tabular}[c]{@{}c@{}}Logistic\\ Regression\end{tabular}} & \textbf{\begin{tabular}[c]{@{}c@{}}Gradient\\ Boosting\end{tabular}} & \textbf{XGBoost} \\ \hline 
\textbf{Accuracy}                                                                    & 0.55                                                             & 0.57                                                                   & 0.58                                                                 & 0.59             \\ \hline 
\textbf{Precision}                                                                   & 0.45                                                             & 0.42                                                                   & 0.47                                                                 & 0.50             \\ \hline 
\textbf{Recall}                                                                      & 0.50                                                             & 0.16                                                                   & 0.32                                                                 & 0.47             \\ \hline 
\textbf{F1-Score}                                                                    & 0.47                                                             & 0.23                                                                   & 0.38                                                                 & 0.48             \\ \hline 
\multicolumn{1}{|c|}{\textbf{\begin{tabular}[c]{@{}c@{}}ROC AUC\\ Score\end{tabular}}} & 0.59& 0.56& 0.59                                                                 & 0.61            
 \\ \hline\end{tabular}
\caption{Test data results from different classifiers}
\end{table}

\begin{figure}[H]
    \centering
    \includegraphics[width=1\linewidth]{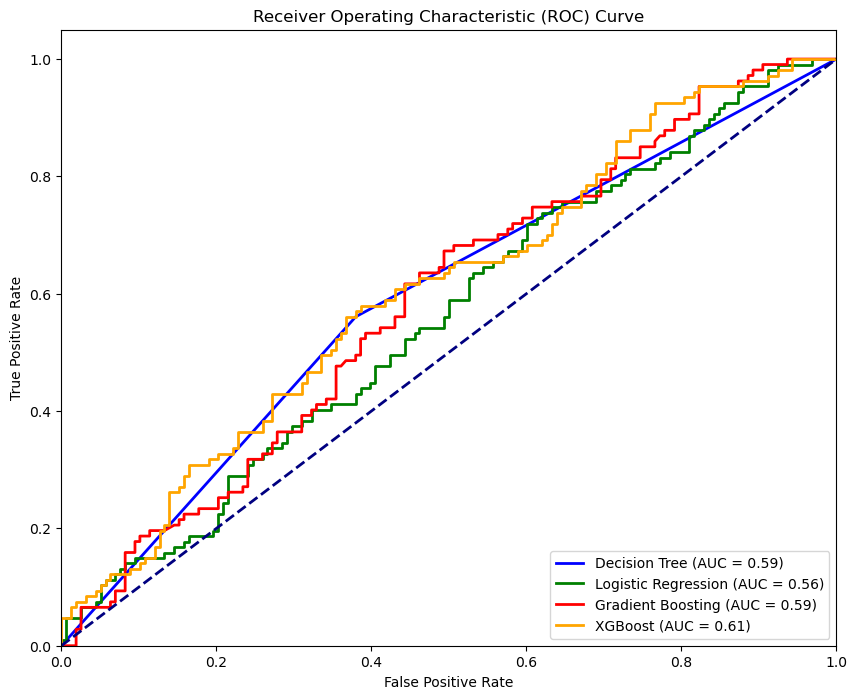}
    \caption{ROC Curve from Different Classifiers}
    \label{fig:enter-label}
\end{figure}

\noindent Table 1. and Figure 3. showing the classification results of two traditional classifiers and two ensemble classifiers. Overall, the ensemble methods outperform traditional classifiers. Consequently, we opt to employ the ensemble models for the next experiments.

\subsection{Including Possession Values and Spatial Data}

\noindent For this experiment, we introduced three more features, namely vaep offensive and 360 features: teammate count and opponent count to enrich our small dataset of 957 instances. Following an 80-20 split for training and testing samples, we compared classification results of the two models which is shown in table 2 and figure 4.

\begin{table}[!h]

\label{tab:my_table}
\begin{tabular}{|l|c|c|} \hline 
                                                                                     & \textbf{\begin{tabular}[c]{@{}c@{}}Gradient Boosting\\ \end{tabular}} & \textbf{XGBoost} \\ \hline 
\textbf{Accuracy}                                                                    & 0.76                                                                 & 0.79             \\ \hline 
\textbf{Precision}                                                                   & 0.65                                                                 & 0.71             \\ \hline 
\textbf{Recall}                                                                      & 0.76                                                                 & 0.75             \\ \hline 
\textbf{F1-Score}                                                                    & 0.70                                                                 & 0.73             \\ \hline 
\multicolumn{1}{|c|}{\textbf{ROC AUC Score}} & 0.82                                                                 & 0.84            
 \\ \hline\end{tabular}
 \caption{Test set results from two ensemble classifier with additional features}
\end{table}
\begin{figure}[H]
    \centering
    \includegraphics[width=1\linewidth]{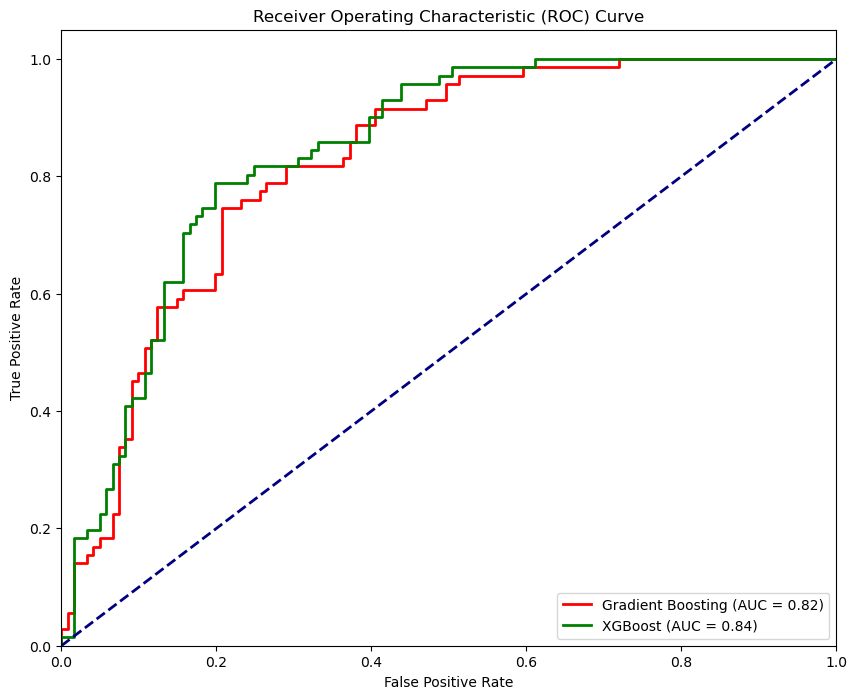}
    \caption{ROC Curve from Gradient Boosting \& XGBoost}
\end{figure}

\begin{table*}[t]
\centering
\begin{tabular}{|c|c|c|c|c|c|}
\hline
\textbf{Competition Name} & \textbf{Season Name} & \textbf{Total Matches} & \textbf{Fouls Committed} & \textbf{Teams Count} & \textbf{Players Count} \\
\hline
Serie A & 2015/2016 & 380 & 2716 & 20 & 446 \\
\hline
Premier League & 2015/2016 & 380 & 2418 & 20 & 402 \\
\hline
La Liga & 2015/2016 & 380 & 3388 & 20 & 452 \\
\hline
Ligue 1 & 2015/2016 & 376 & 2321 & 20 & 424 \\
\hline
Bundesliga & 2015/2016 & 304 & 2013 & 18 & 370 \\
\hline
Indian Super league & 2021/2022 & 114 & 619 & 11 & 195 \\
\hline
FIFA World Cup & 2018 & 64 & 385 & 32 & 246 \\
\hline
FIFA World Cup & 2022 & 64 & 370 & 32 & 253 \\
\hline
UEFA Euro & 2020 & 51 & 329 & 24 & 207 \\
\hline
\end{tabular}
\caption{Training data for the experiment 4.3. [Fouls committed count is after filtering]}
\end{table*}



\noindent We observe a significant improvement in outcomes in terms of accuracy, precision, recall, F1-score, and ROC AUC score following the inclusion of three pivotal features into the dataset. Notably, the accuracy demonstrates an impressive surge of 20\%, while the ROC AUC score experiences a substantial increase of 23\% with XGBoost classifier.
\subsection{Exceptional Performance with Significant Data}

\noindent The publicly available 360 data by StatsBomb covers only 4 competitions and 210 number of matches. Apart from the 360 data, we try to leverage the large amount of traditional event data by StatsBomb hoping to improve our model by using a larger sample space but sacrificing the context provided by the 360 features.

\noindent \\ The data used for training the model include all of publicly released men's event data by StatsBomb at the writing of this paper. Table 3 shows data used to train the model. Apart from this there were few more matches in the StatsBomb data which we utilized for training. We trained the model on a total of 2690 matches with approx 20000 foul events.

\begin{table}[h]
    \centering
    \begin{tabular}{|c|c|}
        \hline
        \textbf{Metric} & \textbf{Value} \\
        \hline
        Accuracy & 0.83 \\
        Precision & 0.83 \\
        Recall & 0.78 \\
        F1-Score & 0.82 \\
        ROC AUC Score & 0.91 \\
        \hline
    \end{tabular}
    \caption{XGBoost Metrics}
    \label{tab:xgboost_metrics}
\end{table}

\begin{figure}[H]
    \centering
    \includegraphics[width=1\linewidth]{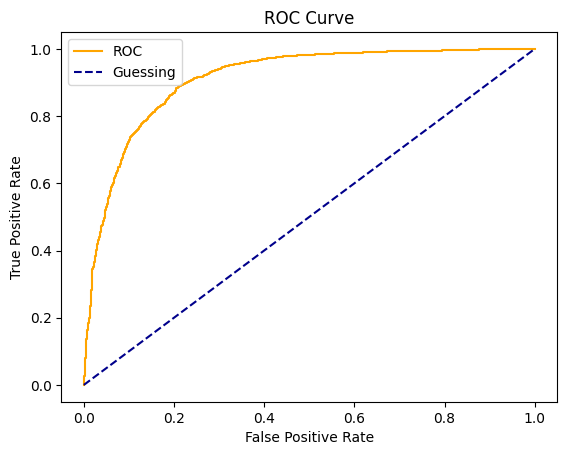}
    \caption{ROC Curve for Experiment 4.3}
    \label{fig:enter-label}
\end{figure}

\begin{figure}[H]
    \centering
    \includegraphics[width=0.8\linewidth]{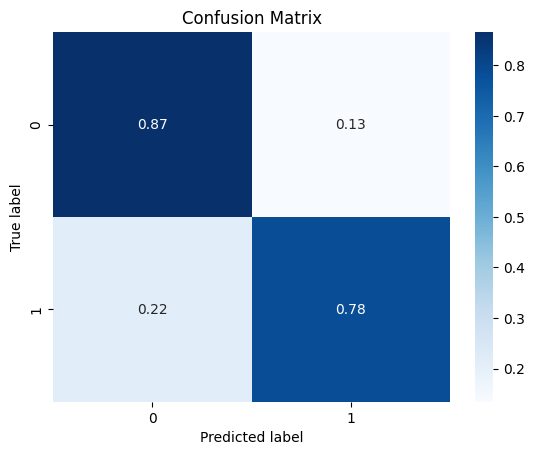}
    \caption{Confusion Matrix for Experiment 4.3}
    \label{fig:enter-label}
\end{figure}

\noindent In this experiment, we aimed to boost model performance by leveraging a substantial amount of traditional event data from StatsBomb, complementing the limited 360 data coverage. The expanded dataset showcased remarkable results for the XGBoost classifier. With an accuracy of 83\%, precision of 83\%, recall of 78\%, F1-score of 82\%, and an impressive ROC AUC score of 91\%, this experiment underscores the effectiveness of utilizing a larger sample space.

\section{ANALYSIS}
\noindent Utilizing the cumulative Expected Booking probability, we can effectively visualize player and team statistics. This section showcases the efficacy of our model by applying it to the FIFA World Cup 2022 data, and presenting the results through comprehensive tables and visualizations. These graphical representations aid in the identification of teams' fouling tactics.

\subsection{FIFA World Cup 2022: Team Statistics}

\noindent \textit{A. Expected Booking (per match) vs Actual Bookings (per match)}

\noindent \\ The y-axis represents the sum of Expected Booking (xB), which is the cumulative probability of receiving a yellow card, divided by the total matches played by the team. The x-axis, on the other hand, represents the sum of actual yellow cards received, divided by the total matches played by the team.

\begin{figure}[H]
    \centering
    \includegraphics[width=1\linewidth]{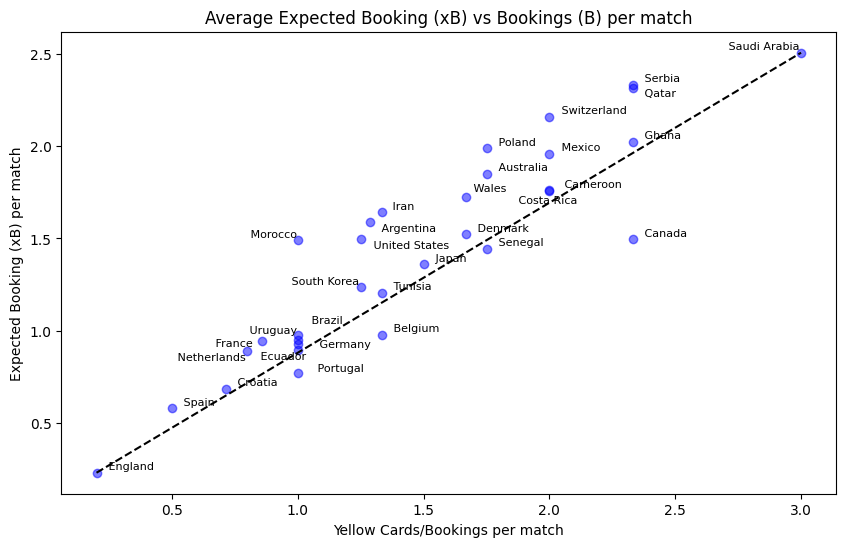}
    \caption{FIFA World Cup 2022 (Yellow card cumulative probability vs Actual yellow card per match)}
\end{figure}

\noindent \textit{B. Ratio (xB/B) vs Fouls per Match}

\noindent \\ The y-axis in this case represents the ratio of the cumulative yellow card probability (xB) to the actual number of yellow cards (B) received. Here, xB stands for expected booking and B stands for actual bookings. A higher ratio indicates a better fouling value. The x-axis represents the number of fouls per match, excluding those that are non-dangerous and not due to bad behaviour.

\begin{figure}[H]
    \centering
    \includegraphics[width=1\linewidth]{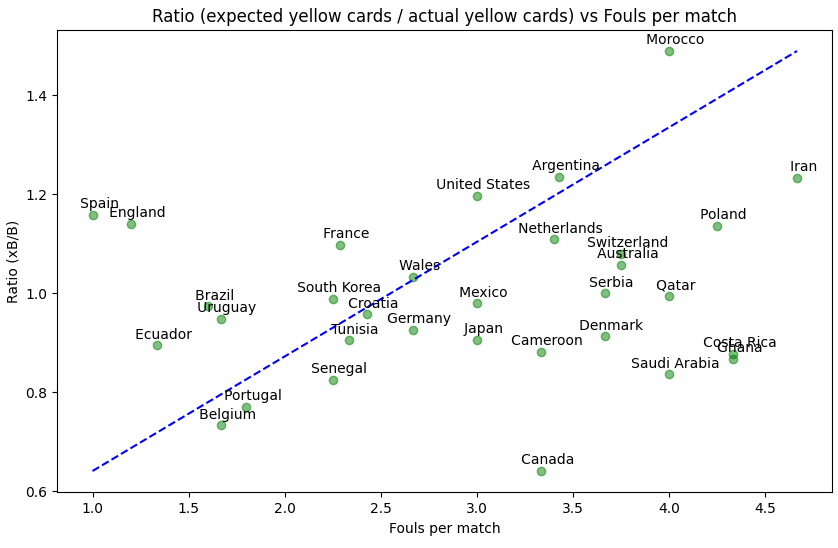}
    \caption{FIFA World Cup 2022 (Ratio vs Fouls per match)}
\end{figure}

\noindent Our analysis reveals that Morocco demonstrated exceptional fouling tactics. They committed more fouls on average in each match, and also successfully avoided yellow cards, as indicated by their high ratio of cumulative yellow card probability to the actual number of yellow cards received.

\subsection{FIFA World Cup 2022: Player Statistics}

\noindent The following plot is generated for players who played a total of at least 90 minutes during the world cup.

\noindent \\ \textit{A. Ratio (xB/B) vs Fouls per 90}

\begin{figure}[H]
    \centering
    \includegraphics[width=1\linewidth]{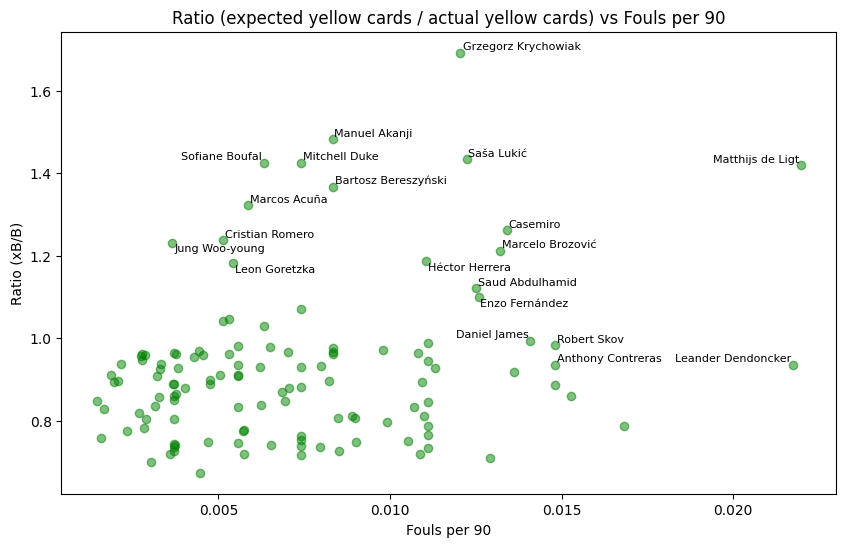}
    \caption{FIFA World Cup 2022 (Ratio vs Fouls per match)}
\end{figure}

\noindent \textit{B. Expected Booking: Best Statistics
}
\begin{table}[H]
\centering
\begin{tabular}{lrrr}
\toprule
Player & xB & B & Ratio \\
\midrule
Marcos Javier Acuña & 3.78& 3 & 1.26\\
Woo-Young Jung & 2.42& 2 & 1.21\\
Fabian Lukas Schär & 1.87& 2 & 0.94\\
Nemanja Gudelj & 1.87& 2 & 0.94\\
Matty Cash & 1.87& 2 & 0.93\\
Jackson Irvine & 1.86& 2 & 0.93\\
Miloš Degenek & 1.79& 2 & 0.89\\
Ngoran Suiru Fai Collins & 1.77& 2 & 0.89\\
\bottomrule
\end{tabular}
\caption{Statistics of players' xB, B, and Ratio.}
\end{table}

\noindent According to our model, Marcos Acuña, the Argentine defender, emerged as the most strategic fouler during the FIFA World Cup 2022.

\section{CONCLUSION}
\noindent In this paper, we present a simple yet novel metric called the Expected Booking (xB) model, designed to estimate the likelihood of a foul resulting in a yellow card. Through three iterative experiments, the model, favoring ensemble methods, exhibited performance improvements with the inclusion of additional features and an expanded dataset. Notably, the analysis of FIFA World Cup 2022 data, which aligns with actual defensive performance, validated the model's efficacy in providing concise insights into team and player fouling tactics. This underscores its ability to provide valuable insights into football events.

\noindent \\ The xB model serves the purpose of addressing a notable gap in the examination of fouling efficiency, shedding light on aspects of defensive strategies often overlooked by prevalent football studies. We assert that further enhancements in the xB model can be attained through the incorporation of more comprehensive data and better spatial features using event freeze frames or tracking data.


{\small
\bibliographystyle{ieee_fullname}
\bibliography{egbib}
}
\end{document}